\documentclass{article}
\usepackage[pdftex]{graphicx}
\usepackage{tikz}
\usepackage{comment}
\usepackage{amsmath,amssymb} % define this before the line numbering.
\usepackage{color}
\usepackage{authblk}
\usepackage{url}

\usepackage{bbding}
\usepackage{pifont}
\usepackage{wasysym}
\DeclareMathOperator*{\argmax}{arg\,max}
\DeclareMathOperator*{\argmin}{arg\,min}
\usepackage{algorithmic}
\usepackage{algorithm}

\usepackage[accsupp]{axessibility}  % Improves PDF readability for those with disabilities.

\usepackage[width=170mm,paperheight=220mm,top=20mm]{geometry}
\begin{document}

\title{Goldilocks-curriculum Domain Randomization and Fractal Perlin Noise with Application to Sim2Real Pneumonia Lesion Detection} % Replace with your title

\author[1]{Takahiro Suzuki}
\author[2]{ Shouhei Hanaoka}
\author[1,2]{Issei Sato}
\affil[1]{Department of Computer Science, Graduate School of Information Science and Technology, The University of Tokyo.} 
\affil[2]{Department of Radiology, The University of Tokyo Hospital.
}

%******************
\date{}
\maketitle

\begin{abstract}
A computer-aided detection (CAD) system based on machine learning is expected to assist radiologists in making a diagnosis. 
It is desirable to build CAD systems for the various types of diseases accumulating daily in a hospital.
An obstacle in developing a CAD system for a disease is that the number of medical images is typically too small to improve the performance of the machine learning model.
In this paper, we aim to explore ways to address this problem through a sim2real transfer approach in medical image fields.
To build a platform to evaluate the performance of sim2real transfer methods in the field of medical imaging, we construct a benchmark dataset that consists of $101$ chest X-images with difficult-to-identify pneumonia lesions judged by an experienced radiologist and a simulator based on fractal Perlin noise and the X-ray principle for generating pseudo pneumonia lesions.
We then develop a novel domain randomization method, called Goldilocks-curriculum domain randomization (GDR) and evaluate our method in this platform.
\end{abstract}

\section{Introduction}

With the recent breakthroughs in deep learning, there have been a growing number of studies on its application to various medical fields \cite{Unet,lungNoduleDetection,NIH,gan_medical,sato}.
As a practical application of machine learning to medical image analysis, it is realistic to expect machine learning models to support experts in making diagnoses, rather than replace experts with models.
We have developed a support system referred to computer-assisted
detection (CAD) system which, after reading clinical images, identifies lesion parts automatically and displays them to a radiologist.
For a multi-institutional study, the CAD system has been in practical use for several diseases since September 2011.

When we develop CAD systems for the various types of diseases accumulating daily in a hospital,
the number of medical images used for training deep neural networks is typically too small to improve their performance.
This paper explores the application of sim2real transfer in the field of medical imaging as a promising future approach to this problem.
Sim2real transfer has been attracting attention in computer vision and robotics as a promising tool to compensate for the lack of annotated data. For instance, in a variety of computer vision tasks, such as object detection \cite{UniformDR,VADRA}, depth estimation \cite{VADRA}, gaze estimation \cite{gaze_estimator,SimGAN}, and semantic segmentation \cite{learnToSimulate}, there have been several studies on training machine learning models on the synthetic data generated by simulators. Similarly, it is also common to train models in the environments created with simulators for robotics control tasks \cite{DRwithBO,UniformDR,simopt,ADR,automateDR,VADRA,deceptionNet}. 

Simulators generally have parameters to determine the distribution for data generation. 
If we successfully set the parameters of a well-designed simulator, natural and realistic images should be generated.
For most cases, however, a simulator will generate poor quality images due to the inappropriate parameters or unsophisticated design of the simulator. 
This gap between synthetic and real data, which is called the sim2real gap, often makes it challenging to transfer the model trained on synthetic data to real-world application. 
A promising approach to close this gap is domain randomization (DR) \cite{UniformDR,learnToSimulate,DRwithBO}, where we make a wide variety of synthetic data by using simulation and train a machine learning model so that it works across all of them. 
If a wide variety of synthetic data contains similar properties to real-world data, training on such synthetic data can improve the model performance in real-world applications.

In this paper, we aim to develop a platform to explore the usefulness of sim2real transfer in the field of medical imaging.
We construct a benchmark dataset that consists of $101$ chest X-images with difficult-to-identify pneumonia lesions judged by an experienced radiologist and a simulator using fractal Perlin noise \cite{fractalPerlinNoise} and the X-ray principle called the Beer-Lambert law \cite{BeerLambert} to simulate the pneumonia lesions.
Although sim2real transfer is expected to be applied to rare diseases with a small number of cases in hospitals, it is generally difficult to publicly release medical images on such diseases for various regulations in hospitals. Therefore, we consider that a first step to evaluate the performance of sim2real transfer is to construct a benchmark dataset for cases with specific characteristics from publicly available data that anyone can access. 

We focus on medical images that is already publicly released by Shin et al. \cite{RSNA}, which was used for the RSNA 2018 Machine Learning Challenge\footnote{\url{https://www.kaggle.com/c/rsna-pneumonia-detection-challenge/}}. 
Although it is usually possible to collect a relatively large number of lesion images in common cases such as pneumonia, we here focus on lesions that are difficult for radiologists, especially inexperienced radiologists, to identify. 
This allows us to publish a benchmark dataset for the application of DR to the small number of medical images with specific characteristics, which is widely accessible to a wider researcher\footnote{We are currently preparing to release our benchmark dataset and simulator}.

We also propose a novel DR method, called Goldilocks-curriculum domain randomization (GDR), in which we train a model continually with the curriculum based on the Goldilocks principle \cite{Goldilocks,85}. 
\begin{quote}
The Goldilocks principle is named by analogy to the children's story ``The Three Bears'', in which a young girl named Goldilocks tastes three different bowls of porridge and finds she prefers porridge that is neither too hot nor too cold, but has just the right temperature.
\hfill Wikipedia
\end{quote}
Unlike most DR methods, we design the DR formulation so that machine-learning models are continuously trained since the original purpose of DR is to train models by using all of the wide variety of synthetic data as we previously describe.
With different simulation parameters, we can create a variety of data at our disposal; thus, when designing the curriculum, we are interested in what type of synthetic data should be used in what order to train the model most efficiently.
We assume that there is an appropriate level of task difficulty in the curriculum as with the right temperature in ``The Three Bears''.

In short, our contributions can be summarized as follows.
\begin{itemize}
    \item We construct a benchmark dataset that consists of 101 chest X-images with specific characteristics, i.e., difficult-to-identify pneumonia lesions (Sec.\,\ref{dataset}).
    \item We construct a simulator based on the fractal Perlin noise and the X-ray principle to generate pseudo-pneumonia images (Sec.\,\ref{simulator}). 
    \item We propose GDR, which is a novel DR algorithm for improving sim2real transfer (Sec.\,\ref{GDR}).
    \item We suggest and experimentally analyze the possibility of catastrophic forgetting in GDRs (Secs.\,\ref{cataforget_inGDR}\,\&\,\ref{discussion}).

\end{itemize}

\section{Background and Related Work}

\subsection{Simulations in Medical Imaging}

Since real medical data with annotation is difficult to acquire in many cases, there have been several studies simulating human tissues and lesions in medical imaging \cite{DLA,mammo,vascular,fractalPerlinNoise,BinaryFractalPerlinNoise}. 
In such simulation, fractal structures are considered essential because such structures naturally appear in human tissues and lesions.
For instance, Bliznakova et al. \cite{mammo} simulated fractal textures in mammographic backgrounds for breast phantom.
Montesdeoca and Li \cite{DLA} used a fractal-growing algorithm called diffusion-limited aggregation \cite{DLA_original} to simulate breast lesions. 
Dustler et al. \cite{fractalPerlinNoise,BinaryFractalPerlinNoise} used Perlin noise \cite{Perlin,Perlin_v2} with fractal structures to simulate breast tissues.

Perlin noise is a well-known algorithm to generate realistic structures and textures and can be applied to create fractal structures.
The noise waves generated at different amplitudes and frequencies are combined to create fractal Perlin noise \cite{Perlin,Perlin_v2,PerlinNoiseDA,fractalPerlinNoise,BinaryFractalPerlinNoise}.
With different amplitudes and frequencies, especially by decreasing an amplitude and increasing a frequency, we can create a large variety of noise waves called octaves; thus, by combining them, natural structures and textures can be obtained on local and global scales.

\subsection{Sim2real Transfer}

Domain adaptation (DA) \cite{AISTAT,SimGAN,CDA} is a common approach to reduce the sim2real gap by updating the data-generating distribution to match the target real data distribution.
In DA, however, a sufficient amount of unlabeled real data are assumed to be accessible to capture the target distribution. 
Thus, taking this approach is infeasible when there are not enough data representing the target domain.

Domain randomization (DR) is another promising approach to reduce the sim2real gap by randomizing properties of simulation data. 
As an early work of DR, Tobin et al. \cite{UniformDR} fixed the data-generating distribution for the task of object detection. 
More specifically, by uniformly sampling the values from the fixed distribution, each aspect regarding the domains, e.g., position, shape, and color of objects, were randomized. 

Adapting simulation parameters for controlling the data-generating distribution has been a recent trend.
For instance, Zakharov et al. \cite{deceptionNet} and Khirodkar et al. \cite{VADRA} set min-max formulation as an objective and trained a model in an adversarial manner to obtain more robust model. 
Ruiz et al. \cite{learnToSimulate} and Muratore et al. \cite{DRwithBO} used the bi-level optimization for the simulation parameters so that the performance of a model trained on the synthetic data is maximized for practical application. 
In more detail, Ruiz et al. \cite{learnToSimulate} used reinforcement learning (RL) and Muratore et al \cite{DRwithBO} used Bayesian optimization (BO) to solve the upper-level problem in the formulation.

For solving the Rubik's cube task, Akkaya et al. \cite{automateDR} constructed a curriculum to gradually increase the task difficulty. 
Although the study is similar to ours in terms of DR with a curriculum, there are crucial differences. First, we propose a curriculum with the Goldilocks principle, which is based on the insight from the recent study \cite{85} in online learning.
Second, we focus on detecting pneumonia lesions, a task of supervised learning rather than RL.

Table \ref{comparison} shows the comparison of the proposed method with current DR methods.

\begin{table}[t!]
    \caption{Comparison of the proposed method with current DR methods
} 
    \label{comparison}
    \centering
    \begin{tabular}{c|c|c|c|c|c} \hline
    & \begin{tabular}{c} Ruiz+\\2019~\cite{learnToSimulate}\end{tabular} & \begin{tabular}{c}Muratore+\\2021~\cite{DRwithBO}\end{tabular} & \begin{tabular}{c}Khirodkar+\\2018~\cite{VADRA}\end{tabular} &
    \begin{tabular}{c}Zakharov+\\2019~\cite{deceptionNet}\end{tabular} &
    Ours\\
    \hline\hline
    Formulation   & bilevel& bilevel & min-max& min-max& curriculum
    \\
    \hline
\begin{tabular}{c}Simulator\\Tuning\\Method\end{tabular} & RL & BO & RL & 
\begin{tabular}{c}Gradient \\ Reversal\end{tabular}
& BO  \\
\hline
\begin{tabular}{c}Continual\\Learning\\ Framework\end{tabular} & N/A & N/A   & N/A  & N/A & \begin{tabular}{c}Goldilocks\\Principle\end{tabular}  \\
\hline
\begin{tabular}{c}Simulator\\Construction\end{tabular}  & \begin{tabular}{c}Unreal\\ Engine\end{tabular} & MuJoCo& \begin{tabular}{c}Unreal \\ Engine\end{tabular}    &
\begin{tabular}{c}Encoder\\Decoder\end{tabular} &
\begin{tabular}{c}Perlin Noise \& \\ X-ray Principle\end{tabular} \\
\hline
\begin{tabular}{c}Appliaction\\Example\end{tabular} & \begin{tabular}{c}Car\\ Counting\end{tabular}  & \begin{tabular}{c}Pendulum \\ Swinging\end{tabular} & \begin{tabular}{c}Car \\Detection\end{tabular}    &
\begin{tabular}{c}Pose \\ Estimation\end{tabular}    &
\begin{tabular}{c}Pneumonia \\Detection\end{tabular}\\
\hline
\end{tabular}
\end{table}

\subsection{Goldilocks Principle in Learning Strategies}
In curriculum learning \cite{curriculumLearning}, it is widely considered that training a model in a meaningful order, especially from easy to hard samples, is effective. 
This might imply, on the basis of Goldilocks principle \cite{Goldilocks}, that there is a sweet spot of task difficulty that is effective to successfully advance the training of the current model. 
Regarding this issue, the role of task difficulty on the rate of learning for binary-classification tasks has been examined \cite{85}. 
From a theoretical analysis under certain assumptions, it was found that the optimal error rate for training is about $15$ \%, i.e., the optimal training accuracy is about $85$ $\%$.  
Even if task difficulty is set constant, the model performance has been improved by learning, i.e.,  the next task is more difficult in terms of the previous model performance.

\section{Proposed Method}

In this section, we first describe the simulator we constructed for detecting difficult-to-identify pneumonia lesions in chest X-images. 
We then explain the details of GDR.
Finally, we consider the possibility of catastrophic forgetting in GDR and its solution.

\subsection{Simulator Construction}
\label{simulator}

We construct the simulator that, given a normal chest X-image as an input, outputs an abnormal image where pseudo pneumonia lesions are randomly inserted into the normal lungs. 
To the best of our knowledge, we are the first to exploit fractal Perlin noise to simulate pneumonia lesions. 
We also apply the X-ray principle called the Beer-Lambert law \cite{BeerLambert} as a synthesis method for generating pneumonia lesions.
We describe the details of the procedures for generating pseudo pneumonia lesions and determining where and how to insert the lesions into a normal image in Appendix \ref{simulator_detail}.

\begin{figure}[t!]
\centering
\includegraphics[height=6cm]{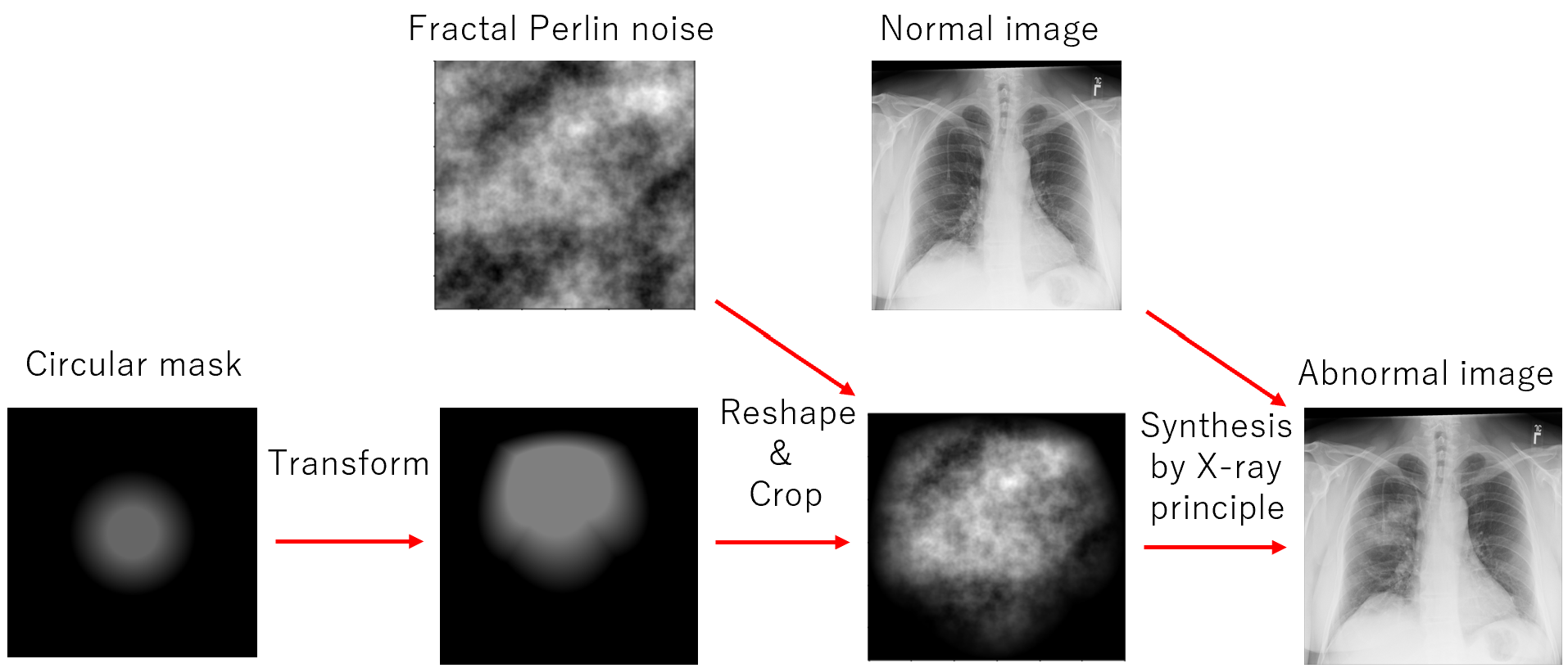}
\caption{Overview of our simulator}
\label{flow}
\end{figure}

The proposed simulator has the parameters 
\begin{itemize}
    \item to control the size of the pseudo lesion,
    \item to control the smoothness around the edges of the lesion,
    \item to control the whiteness of the lesion,
    \item to generate fractal Perlin noises.
\end{itemize}
Note that the value of the parameter related to size does not directly determine the lesion size, because we randomly transform the shape of the generated lesion through an affine transformation to create various shape patterns.
Also, the parameters for smoothness and whiteness are considered essential to generate realistic pneumonia images in our simulator.
The details of these parameters are described in Appendix \ref{smooth} and \ref{white}. 
The parameters for generating fractal Perlin noise are explained specifically in Sec.\,\ref{parameters}. 
The whole procedure of the proposed simulator is illustrated in Fig.\,\ref{flow}.

\subsection{Goldilocks-curriculum Domain Randomization}
\label{GDR}
With GDR, we continually train a model so that the model performs well not only on the current training data but also on all of the previous training data. 
More formally, let $L$ be a loss function and $\theta$ be the model parameter. 
Then, $\theta^t$ is a minimizer of the cumulative loss function at the current timestep $t$, i.e.,
\begin{align}
\label{continual}
    \theta^t = \argmin_{\theta} \sum_{i=1}^t L(\theta, S_{\phi^i}),
\end{align}
where $S_{\phi^i}$ represents the training data at the timestep $i$, generated from the simulator with its parameter value $\phi^i$.

\subsubsection{Goldilocks-curriculum.}
In our continual learning on simulation, we introduce a curriculum to determine what type of training samples should be generated and used in the next timestep.
Inspired by the recent study \cite{85} on analyzing the optimal task difficulty for binary classification in online learning, we develop the Goldilocks curriculum 
that can be applied to DR for object detection tasks.

In our Goldilocks curriculum, we use the simulation data on which the performance of the current model is the target value $k$; that is, $k$ is a hyperparameter. 
More specifically, we find the following simulation parameter $\phi^{t+1}$ and use it to generate synthetic data in the next timestep $t+1$: 
\begin{align}
\label{k-percent}
    \phi^{t+1} \in \{\phi \in \Phi \, | \, V(\theta^t, S_{\phi}) = k\}, 
\end{align}
where $\Phi$ is the simulation parameter space, $V(\theta^t, S_{\phi})$ represents the performance score of the current model with its model parameter ${\theta^t}$, on the synthetic data generated using the simulation parameter $\phi$.
In the formulation \eqref{k-percent}, the hyperparameter $k$ controls the difficulty of the training samples. More specifically, the higher the value of $k$ is set, the easier the training samples will be for the current model. 
In the context of curriculum learning, the Goldilocks principle \cite{Goldilocks} suggests that there might be a sweet spot of task difficulty that is effective to successfully progress the training of the current model.
Deriving the optimal difficulty $k$ is infeasible in object detection based on deep neural networks; thus, we set the value empirically through cross-validation.

Note that $V(\theta^t, S_{\phi})$ is typically a black-box function with respect to $\phi$.
For such a non-differentiable simulator, we instead use the following formulation to search for the simulation parameter that generates the synthetic data on which the performance of the current model is around the target $k$.
\begin{align}
\label{k-percent-Bayes}
    \phi^{t+1} = \argmax_{\phi \in \Phi} \, - | V(\theta^t, S_{\phi}) - k |.
\end{align}
The proxy problem \eqref{k-percent-Bayes} can be effectively solved using a black-box optimization method such as BO. 
The algorithm of GDR is summarized in Algorithm \ref{GDR_algo} in Appendix.

\subsection{Catastrophic Forgetting in GDR}
\label{cataforget_inGDR}

The most naive approach to solve the problem \eqref{continual} is replaying, i.e., using not only the current data $S_{\phi^t}$ but also all or some of the previous data $\{S_{\phi^i}\}_{i=1}^{t-1}$ for training the model.
Yet, it takes substantial computational cost due to the retraining process. 
Since the previous data were already used to train the model, if we can successfully keep the high performance of the model on the previous data, just using the current data is enough for training the model. 
In this paper, we focus on optimization methods to prevent catastrophic forgetting in GDR.
More specifically, optimizers such as NVRM-SGD \cite{VSGD} can lead to finding a flat minima of the loss function, which, according to the analyses in previous studies \cite{hesse,VSGD}, is effective for mitigating catastrophic forgetting.
That is, we can compute the forgetting score at timestep $t$, $L\left(\theta^{\textcolor[rgb]{1,0,0}{t}},S_{\phi^{\textcolor[rgb]{1,0,0}{t-1}}}\right)$, by using a second-order Taylor approximation of the loss surface around the optimum at timestep $t-1$, $\theta^{t-1}$ as follow. 
\begin{align}
L\left(\theta^{t},S_{\phi^{t-1}}\right)
%& \approx L\left(\theta^{t},S_{\phi^t}\right)+\left(\theta^{t+1}-\theta^{t}\right)^{\top} \nabla L\left(\theta^{t},S_{\phi^t}\right)+\frac{1}{2}\left(\theta^{t+1}-\theta^{t}\right)^{\top} \nabla^{2} L\left(\theta^{t},S_{\phi^t}\right)\left(\theta^{t+1}-\theta^{t}\right) \\
& \approx L\left(\theta^{t-1},S_{\phi^{t-1}}\right)+\frac{1}{2}\left(\theta^{t}-\theta^{t-1}\right)^{\top} \nabla^{2} L\left(\theta^{t-1},S_{\phi^{t-1}}\right)\left(\theta^{t}-\theta^{t-1}\right)\label{Eq:flatness_foregetting}
\end{align}
where $\nabla^{2} L\left(\theta^{{t-1}},S_{\phi^{t-1}}\right)$ is the Hessian for $L(\theta,S_{\phi^{t-1}})$ at $\theta^{t-1}$ and positive semi-definite, and note that $\theta^{t-1}$ is assumed to converge to a stationary point of $L(\theta,S_{\phi^{t-1}})$, i.e., $\nabla L\left(\theta^{t-1},S_{\phi^{t-1}}\right) \approx \mathbf{0}$. 
Equation \eqref{Eq:flatness_foregetting} means that reducing the maximum eigenvalue of the Hessian, i.e., the flatness of a loss landscape, leads to less forgetting.
Thus, we propose to use NVRM-SGD as an optimizer in GDR, instead of commonly used optimizers, such as Adam \cite{Adam}, that can easily cause catastrophic forgetting without replaying.

\section{Experiments}
We empirically investigated the effectiveness of GDR with our simulator for detecting difficult-to-identify pneumonia lesions in chest X-images.
We first constructed a dataset for evaluation consisting of $101$ pneumonia images that radiologists found difficult to identify.
We then analyze the proposed method and other baseline methods.
We also demonstrated that using NVRM-SGD\cite{VSGD} is effective to mitigate catastrophic forgetting in GDR by comparing the performance shifts depending on the choice of an optimizer.
Finally, we empirically confirmed Goldilocks effect in GDR.

\subsection{Dataset}
\label{dataset}

We created the dataset comprised of difficult-to-identify pneumonia images in the following procedure. First, from the dataset released by Shin et al. \cite{RSNA}, which was used for the RSNA 2018 Machine Learning Challenge\footnote{\url{https://www.kaggle.com/c/rsna-pneumonia-detection-challenge/}}, we gathered $26684$ chest X-ray images with annotations. From the images, we then collected the chest X-images that satisfy all of the following conditions: (i) view position = ``PA'', (ii) age $\geq$ 18, and (iii) label = ``Lung Opacity'', where ``PA'' means posterior-anterior projection and ``Lung Opacity'' means pneumonia in this context. Through the above filtering, we obtained $1288$ abnormal images with bounding boxes. Furthermore, we focused on the size of bounding boxes on the basis of the fact that smaller lesions tend to be more difficult for radiologists to identify. 

Considering that the size of each image is $1024$ $\times$ $1024$ pixels, the lesion size of $150$ $\times$ $150$ pixels might be considered small. 
Therefore,  we selected $142$ abnormal images containing at least one bounding box the longer side of which is $150$ pixels or less. 
Finally, for each of the $142$ abnormal images, an experienced radiologist judged whether it could be seen as a difficult-to identify lesion through giving a score on a scale of one to ten, and the top $101$ images were selected to compose the dataset. 

Similarly, we also selected the $6881$ normal images that meet the above conditions (i), (ii), and another condition (iv) label = ``Normal''. 
Among the normal images, we randomly used $1000$ images for generating pseudo-rare pneumonia images as training data in the DR methods.

With GDR, we randomly selected $200$ additional normal images to generate abnormal images.
These abnormal images were used for determining an appropriate epoch when training a lesion detection model.

\subsection{Evaluation Metrics}
Following the previous studies on lung nodule detection \cite{lungNoduleDetection,KenjiSuzuki,solitaryLungNodule}, we used two evaluation metrics: area under the free-response receiver operating characteristic curve (FAUC) \cite{FAUC} and competition performance metric (CPM) \cite{CPM}.
FAUC is used to measure the average true positive rate (TPR) in the area specified for the number of false positives (FPs) per image, while CPM represents the average TPR at the following seven points: $\frac{1}{8}, \frac{1}{4}, \frac{1}{2}, 1, 2, 4$, and $8$ FPs per image.
Following the previous studies, we designated the area under one FP per image in the FROC curve as FAUC. 
We also calculated TPR when the number of FPs per image is limited to $0.2$ on average. 
Since excessive FPs can confuse radiologists, we interviewed a radiologist and concluded that a value of $0.2$ is acceptable for practical application.

Both FAUC and CPM metrics can be calculated by drawing the FROC curve. The FROC curve is known as a tool for characterizing the performance of a free-response system at all decision thresholds simultaneously; thus, it is especially useful for the detection and localization tasks in which multiple regions of interests (ROIs) may exist in an input. 
The vertical axis of the FROC curve represents TPR and the horizontal axis represents the average number of FPs per image. Here, TPR means the proportion of ROI detected with the machine learning model. 
When determining the detection condition, we have to consider the application in clinical practice: it is crucial that there be a recognizable overlap between ROI and bounding boxes predicted with the model, because a radiologist will pay attention to the area around the predicted bounding boxes. 
Therefore, in our experiments, ROI is considered detected if the dice coefficient of the ROI and a bounding box predicted with the method is $0.2$ or larger. Similar strategies were used in previous studies on detection tasks \cite{lungNoduleDetection,KenjiSuzuki,solitaryLungNodule}.

The procedure to draw the FROC curve is as follows. 
First, set the list of thresholds for the confidence score of prediction. Next, for each threshold in the list, output the bounding boxes with confidence scores larger than the threshold. Then, determine whether each predicted bounding box is FP, TP, or should be ignored. More specifically, a prediction is considered as FP if it does not match any ground truth. i.e., the dice coefficient between the prediction and any ROI is less than $0.2$. On the other hand, a prediction is regarded as TP if it matches a ground truth. Note that we ignored the prediction that cannot be regarded as correct or incorrect as follows. 
\begin{itemize}
    \item There is a match with a ground truth, but the ground truth has already been detected and matched with another prediction.
    \item There is a match with a ground truth, but its lesion size is not small and not of interest in the experiment.
\end{itemize}
Next, we count the number of TPs and FPs across all input images and calculate FPs per image and TPR at the threshold. Finally, since the FROC points are discrete, we used linear interpolation to draw the FROC curve.

\subsection{Simulator Implementation}
\label{parameters}

We used the distributed package\footnote{\url{https://github.com/pvigier/perlin-numpy}} to generate $2$-dimensional fractal Perlin noise. The function has several arguments to generate such noise. 
Using prior knowledge on pneumonia lesions, we determined a range of several parameters.
For instance, persistence represents how much we decrease amplitudes, while lacunarity means how much we increase frequency. We set the domain of persistence from $0.2$ to $1$ and lacunarity from $2$ to $4$, respectively.
Also the argument res, denoting the number of nose periods, was set to take a discrete value from $2$ to $5$.
We fixed the number of octaves to $5$ because the appearance of generated lesions looked natural.
Another argument shape, which determine the size of the generated noise, does not have any effect on the appearance of the image. Therefore, any valid value is suitable for this parameter; we set $shape:=lacunarity^{(octaves-1)}*res$, following the remarks in the package. 
Moreover, every time generating a lesion, we changed the value of initial seeds used in the package to ensure a diversity of visual appearances. 

The implementation details with regards to size, smoothness, and whiteness are as follows. To provide a variety of lesion sizes, we first did not fix the size parameter $r$. 
Instead, every time we generated a pseudo lesion, we sampled a value of the parameter $r$ from a uniform distribution. 
We fixed the lower-bound of the uniform distribution to $20$ and the upper-bound to $75$ pixels, on the basis of the minimum and maximum sizes of the pneumonia lesions in the $101$ pneumonia dataset.
We also set the domain of the smoothness parameter from $0.2$ to $0.8$ and the whiteness parameter from $0.1$ to $1$, respectively.

Thus, these five parameters, i.e., persistence, lacunarity, res, smoothness, whiteness, are adjustable within their pre-defined ranges respectively. 
In Uniform DR (UDR) \cite{UniformDR}, the values of these parameters are uniformly sampled from the ranges. 
In Baysian DR (BayRn) \cite{DRwithBO}, these parameters are optimized by using BO to maximize the FAUC performance on the validation data. 
For each timestep in the curriculum with GDR, these parameters are adjusted through BO to generate abnormal images of the target difficulty for the current model.

\subsection{BO Settings}
In this section, we explain how to query the next paramters of our simulator.
We use BO which is a framework for an optimization of the financially, computationally or physically expensive black-box functions whose derivatives and convexity properties are unknown.
We focus on  the problem of optimizing a black-box function $f(\phi)$ over a compact set $\Phi\subset \mathbb{R}^d$.
\begin{align}
\phi^*=\argmax_{\phi \in \Phi} f(\phi).
\end{align}
By modeling $f$ as a function distributed from a Gaussian process (GP), nearby locations have close associated values where ``nearby'' is defined by the kernel of the GP.
Finding the maximum of $f$ is achieved by generating successive queries $\phi^{(1)},\phi^{(2)},\ldots \in \Phi$.
The common approach to find the next query is to solve the alternative optimization problem whose objective function is relatively cheap to optimize.
That is, we select the next query by finding the maximum of an {\it acquisition function} $a(\phi)$ over a bounded domain in $\Phi$ instead of finding the maximum point of the objective function $f(\phi)$.

The acquisition function is formulated by the posterior mean and variance conditioned on the GP kernel to balance exploration and exploitation.
In this work, we use the expected improvement (EI),  which is well used acquisition function in BO.
After choosing the acquisition function, we need to also determine the kernel function and its parameter that governs the behavior of the GP.
We use the squared exponential (SE) kernel which is also well used in the GP given by 
\begin{align}
\kappa_\mathrm{SE}(\phi,\phi^{\prime})&=\exp\left(-\frac{\gamma}{2}\sum_{j=1}^{d}(\phi_j-\phi^{\prime}_j)^2\right),
\end{align}
where we set the kernel parameter $\gamma=0.25$.
We can optimize the kernel parameter from observations; however, it needs many BO trials.
We thus fix the kernel parameter.

Intuitively, our setting means that we use information from a relatively wide range of observed data, which assumes smoothness in the underlying black-box function.
It is also assumed in the SE kernel that the black-box function is a relatively smooth. 
The purpose of this work is to obtain a query with a high score in a relatively small number of BO trials, and it cannot be achieved unless the black-box function is assumed to be smoothness.

Since it is difficult to find the maximum solution of the acquisition function,
We use Latin hyper-cube sampling to generate  samples over hyper-cube $[0,1]^d \subset \mathbb{R}^d$
Latin hyper-cube sampling prevents the sample points from clumping together in sample space, which can happen with purely random points.

For each parameter, i.e., persistence, lacunarity, res, smoothness, we apply a linear transformation to map the parameter's range to $[0,1]$. 
In more detail, let $x \in [l,u]$ be the parameter, where $l$ and $u$ represent the lower and upper bound, respectively.  
Then, it is normalized to $y \in [0,1]$ given by $y = \frac{x-l}{u-l}$.
For the opposite transformation, we have $x = l + (u-l)y$.
When the parameter $x$ should be discrete, we integerized using the floor function: $x = \lfloor l + (u-l+1)y \rfloor$.

\subsection{Experimental Settings}
For each DR method, we randomly selected $1000$ normal images to generate abnormal images with our simulator as training data.  
We also applied $5$-fold cross-validation to evaluate the performance of each method. 
More specifically, $20$ abnormal images from the $101$ pneumonia dataset and the same number of randomly-selected normal images were used for validation, while the remaining $81$ abnormal images in the dataset and the same number of normal images were used for testing.

Furthermore, we used a single shot multibox detector (SSD) \cite{ssd} as a detection model for pneumonia lesions. 
The model was pretrained on the ImageNet dataset \cite{imagenet} and provided by the torchvision package ver.\,1.10.0. 
We downsampled each image to the size of $300$ $\times$ $300$ pixels, which is the expected input size of the SSD model. 
As optimizers for training, We used Adam \cite{Adam}, SGD, and NVRM-SGD \cite{VSGD}. Regarding the argument about the neural variability scale in NVRM-SGD, we used the default value of $0.01$.
We also used SAM \cite{SAM} in GDR with the argument about the neighborhood size $0.05$.
We all set their learning rates to $0.0002$.
For each training, we set the batch size to $64$ and the number of epochs to $120$.
The implementation details of each method are described below.

\subsubsection{BayRn \cite{DRwithBO}}
We optimized the five parameters in our simulator, i.e., persistence, lacunarity, res, $\alpha$, $\beta$, to maximize the validation performance of the model trained on the synthetic data.
More formally, let $S_{\rm{val}}$ be the validation dataset. Then, we solve the following bilevel problem to optimize the simulation parameter $\phi$:
\begin{align}
\label{upper}
\phi^* &= \argmax_{\phi \in \Phi} V(\theta^*(\phi), S_{\rm{val}})\\
\label{lower}  \rm{s.t.} ~~ \theta^*(\phi) &= \argmin_{\theta} L(\theta, S_{\phi}).
\end{align}
The lower-level problem \eqref{lower} represents the training of the model on the synthetic data $S_{\phi}$. 
The upper-level problem \eqref{upper} corresponds to optimizing $\phi$ to maximize the validation performance of the model trained with the lower-level problem \eqref{lower}. We used FAUC as a performance metric $V$ in \eqref{upper}.
In BayRn, BO is used to solve the problem \eqref{upper}: we set the number of initial seeds to $5$ and the number of total iterations to $40$ in BO.

\subsubsection{GDR}
We randomly selected $200$ normal images to generate abnormal images with our simulator as validation, i.e., to determine an appropriate epoch when training the model.
Note that in GDR, the validation data comprised of the normal and abnormal X-ray images were used to determine which curriculum is the best, serving as a role of meta-validation: they were used for determining an appropriate target score and iteration in the curriculum.

Regarding the curriculum, we used FAUC as a target metric and tried the following target performance scores: 0.3, 0.4, 0.5, 0.6, and 0.7 FAUC. 
For each target score, the parameters of our simulator were adjusted through BO to generate abnormal images that achieve the target performance score for the current model. 
In the BO process, we set the number of initial seeds to $5$ and the number of total iterations to $35$. Note that, to reduce rendering time, we used just $30$ out of the $1000$ normal images to generate abnormal images when finding the target parameters of the simulator.
For each target score, we continually trained the detection model up to $10$ timesteps in the curriculum.

\subsubsection{Easy2Hard-curriculum}
To confirm the effectiveness of Goldilocks-curriculum in GDR, we tried different curriculum to decide the next simulation parameter in the expression \eqref{k-percent}.
Several pacing functions have been proposed for curriculum learning; however, since there is not much performance difference among functions \cite{Curricula}, we used step-pacing functions as the curriculum to be compared in this experiment for easy interpretation of pace.

\section{Results and Discussions}
\label{discussion}

Table \ref{performances} shows the test performances of the different methods on the $101$ pneumonia dataset.  
We observed that GDR achieved the best performances compared with baseline methods in terms of FAUC, CPM, and TPR at the threshold with which the number of FPs per image is $0.2$. 
We also confirmed that Goldilocks-curriculum performed better than the easy2hard curricula.
These results indicate that GDR might be a promising DR approach. 
See Fig.\,\ref{FROC} in Appendix for FROC curves for each method.

The reason why GDR outperformed other DR methods might be as follows. 
Since the data-generating distribution is fixed and not optimized in UDR, unrealistic abnormal images tend to be generated, hindering to improve the model performance. 
It is considered that relatively a large number of images are necessary in UDR to ensure the sufficient amount of realistic abnormal images. 
In BayRn, since the simulation parameters were optimized to maximize the performance on the validation data, the model trained on the synthetic data might fail to capture the features of the pneumonia lesions that were not appeared in the validation data. 
That is, this approach seems to be ineffective for the tasks in which the number of real data in the validation is too small to be representative of the test data.
On the other hand, we continuously train a model in GDR; by exposing the model to a more variety of synthetic data, the model would be more robust and the performances would be much improved.

\setlength{\tabcolsep}{4pt}
\begin{table}[t!]
\begin{center}
\caption{Performances of different methods on $101$ pneumonia dataset. \textbf{Easy2Hard-1 and 2} are formulated by using curriculums with step-pacing functions ``$1 \rightarrow 0.9\rightarrow \cdots 0.2 \rightarrow 0.1$'' and ``$0.75 \rightarrow 0.7 \rightarrow\cdots 0.25 \rightarrow 0.3$'', respectively. \textbf{TPR@FPI0.2} is a TPR whose FPs/Image is $0.2$.}
\label{performances}
\begin{tabular}{llll}
\hline\noalign{\smallskip}
Method  & FAUC & CPM & TPR@FPI$0.2$\\
\noalign{\smallskip}
\hline
\noalign{\smallskip}
    UDR (Adam) & 0.324 $\pm$ 0.037 & 0.394 $\pm$ 0.026 & 0.245 $\pm$ 0.040\\
    UDR (SGD) & 0.257 $\pm$ 0.021 & 0.335 $\pm$ 0.032 & 0.197 $\pm$ 0.011\\
    UDR (NVRM-SGD) & 0.246 $\pm$ 0.023 & 0.361 $\pm$ 0.030 & 0.175 $\pm$ 0.029\\
    BayRn (Adam) & 0.320 $\pm$ 0.035 & 0.410 $\pm$ 0.044 & 0.251 $\pm$ 0.056\\
    BayRn (SGD) & 0.302 $\pm$ 0.044 & 0.388 $\pm$ 0.049 & 0.232 $\pm$ 0.038\\
    BayRn (NVRM-SGD) & 0.282 $\pm$ 0.026 & 0.384 $\pm$ 0.028 & 0.203 $\pm$ 0.028\\
    GDR (Adam) & 0.352 $\pm$ 0.028 & 0.452 $\pm$ 0.027 & 0.275 $\pm$ 0.039\\     
    GDR (SGD) & 0.377 $\pm$ 0.032 & 0.474 $\pm$ 0.036 & 0.302 $\pm$ 0.035\\
    GDR (NVRM-SGD) & $\mathbf {0.378 \pm 0.015}$ & $\mathbf {0.491 \pm 0.009}$ & $\mathbf {0.331 \pm 0.034}$\\
    GDR (SAM) & 0.354 $\pm$ 0.033 & 0.471 $\pm$ 0.022 & 0.291 $\pm$ 0.046\\     
    Easy2Hard-1 (SGD) & 0.350 $\pm$ 0.033 & 0.457 $\pm$ 0.049 & 0.276 $\pm$ 0.050\\
    Easy2Hard-1 (NVRM-SGD) & 0.362 $\pm$ 0.024 & 0.470 $\pm$ 0.016 & 0.297 $\pm$ 0.038\\
    Easy2Hard-2  (SGD) & 0.365 $\pm$ 0.017 & 0.459 $\pm$ 0.032 & 0.289 $\pm$ 0.015\\
    Easy2Hard-2 (NVRM-SGD) & 0.363 $\pm$ 0.044 & 0.466 $\pm$ 0.046 & 0.320 $\pm$ 0.054\\
\hline
\end{tabular}
\end{center}
\end{table}
\setlength{\tabcolsep}{1.4pt}

\begin{figure}[t!]
\centering
$
\begin{array}{cc}
\includegraphics[width=8cm]{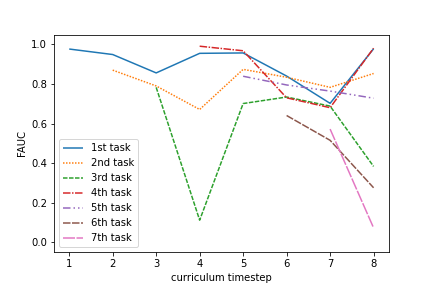}&
\includegraphics[width=8cm]{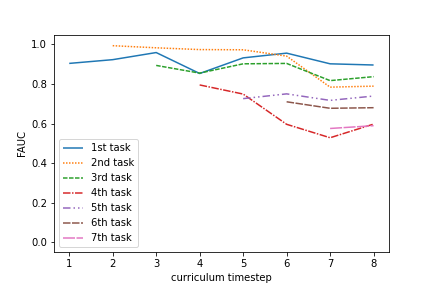}
\\
({\rm a})~{\rm Optimizer: Adam} 
&
({\rm b})~{\rm Optimizer: SGD} 
\\
\includegraphics[width=8cm]{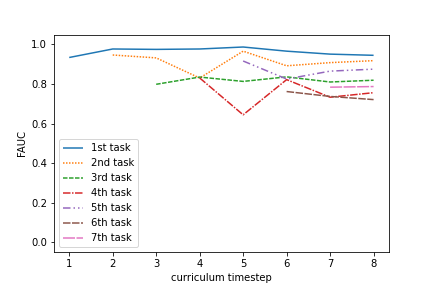}&
\includegraphics[width=8cm]{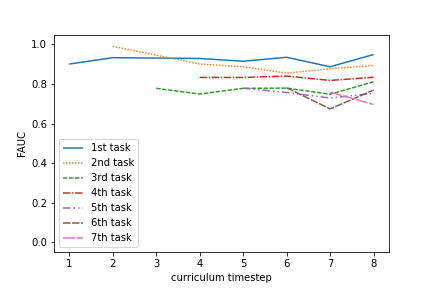}
\\
({\rm c})~{\rm Optimizer: SAM} 
&
({\rm d})~{\rm Optimizer: NVRM-SGD} 
\end{array}
$
\caption{Performance shift relative to number of timesteps with different optimizers. Adam (left top), SGD (right top), SAM (left bottom), and NVRM-SGD (right top) were used as optimizers, respectively. The target performance score was set to $0.6$ FAUC.}
\label{cataforgetfig}
\end{figure}

We then investigated how the choice of an optimizer would affect catastrophic forgetting in GDR. 
Figure \ref{cataforgetfig} shows how the performance shifts as the number of timesteps of the curriculum increases.
We used Adam, SGD, SAM, and NVRM-SGD as optimizers. 
Catastrophic forgetting apparently occurred when Adam was used. 
In contrast, catastrophic forgetting was successfully prevented by using NVRM-SGD.
We also confirmed that SGD can mitigate catastrophic forgetting compared with Adam, which is consistent with the recent study \cite{hesse}.
SAM can mitigate catastrophic forgetting to certain degree.
These results and the performances in Table \ref{performances} might support our hypothesis that the model performances can be improved by preventing catastrophic forgetting in GDR.

Table \ref{performances} shows that GDR with NVRM-SGD outperforms other settings and baseline methods on all evaluation metrics.
In particular, GDR with NVRM-SGD achieved around $0.33$ TPR on average under the condition that the number of FPs per image is limited to $0.2$, which we call TPR@FPI$0.2$, and is much higher than other methods in this evaluation.
This means that we will be able to detect about one-third of the difficult-to identify pneumonia lesions in chest X-rays by using the method. 
Since these lesions are difficult for radiologists to identify, detecting even a third of them is considered useful as a practical CAD system.

We also compared GDR with two Easy2Hard curricula that are formulated by using two step-pacing functions. Easy2Hard-1 is a commonly used step-pacing curriculum; however, its performance is lower than that of the GDR. 
To further analyze these results, we consider Easy2Hard2-2 that changes near a specific difficulty level and has a narrower step width, which is considered to be similar in nature to the GDR.
As a result, Easy2Hard-2 has better performance than Easy2Hard-1.
This suggests that that curriculum learning may be more effective when the difficulty is fixed at a specific level.

We also confirmed the Goldilocks effect in our setting. 
Table \ref{goldilocks} shows how the performance changes depending on the target performance score with GDR.
Note that we focused on using NVRM-SGD as an optimizer since the performances was superior as shown in Table \ref{performances}.
The target score around $0.6$ FAUC is the optimal difficulty for training the model for this setting. 
As the task difficulty increases (i.e., lower target FAUC) or decreases (i.e., higher target FAUC), 
the performance of the model trained on the task degraded.

\setlength{\tabcolsep}{4pt}
\begin{table}[t!]
\begin{center}
\caption{Effect of target performance score in GDR where NVRM-SGD is used as an optimizer}
\label{goldilocks}
\begin{tabular}{llll}
\hline\noalign{\smallskip}
Target & FAUC & CPM & TPR at FPs/Image=0.2\\
\noalign{\smallskip}
\hline
\noalign{\smallskip}
    FAUC=0.3 & 0.335 $\pm$ 0.047 & 0.444 $\pm$ 0.047 & 0.277 $\pm$ 0.058\\
    FAUC=0.4  & 0.344 $\pm$ 0.040 & 0.457 $\pm$ 0.039 & 0.307 $\pm$ 0.048\\      
    FAUC=0.5  & 0.365 $\pm$ 0.025 & 0.481 $\pm$ 0.020 & 0.302 $\pm$ 0.026\\      
    FAUC=0.6 & $\mathbf {0.377 \pm 0.015}$ & $\mathbf {0.483 \pm 0.017}$ & $\mathbf {0.312 \pm 0.034}$\\
    FAUC=0.7  & 0.357 $\pm$ 0.021 & 0.471 $\pm$ 0.019 & 0.283 $\pm$ 0.035\\      
\hline
\end{tabular}
\end{center}
\end{table}
\setlength{\tabcolsep}{1.4pt}

\section{Conclusion}
We first constructed a simulator based on the fractal Perlin noise and the X-ray principle to generate pseudo rare pneumonia images.
We also constructed a benchmark dataset that consists of $101$ chest X-images including rare pneumonia lesions judged by an experienced radiologist for evaluation.
Our empirical results indicate that our method, GDR with our simulator, performed better than the baseline methods for the detection of the difficult-to identify  pneumonia lesions in chest X-images.
We empirically found the phenomenon of catastrophic forgetting when training a model continually in DR and showed using an optimizer to prevent catastrophic forgetting might be essential to improve model performance.
Since GDR is a general framework and can be applied to a wider range of tasks beyond pneumonia detection, it can be a promising approach for cases where only a small amount of real annotation data is available but a simulator is available. 

\begin{paragraph}{\textbf{Limitation and applicability.}}
Our simulator is limited in terms of the diseases to which they can be applied. Fractal perlin noise, as evidenced by its common use in computer graphics is to give objects a natural appearance, is well suited for representing natural textures. 
That is, it can simulate the natural structure of tissues in the human body as lesions. 
Conversely, it is not suitable for simulating unnatural morphological abnormalities that grow by destroying fractal structures such as cancer.

We consider that our simulator can be applied to the following lesions.
Irregular and obscure margin often occurs when an pathological entity,  e.g., malignant tumor, inflammation, and infection, spread  in-homogeneously. Not only pneumonia but many pathological entities often show such infiltrative, scatter, or obscure margin in its progress such as brain tumor with in-homogeneous  damage of blood brain barrier (BBB).
\end{paragraph}

%\clearpage
%\bibliographystyle{splncs04nat}
\bibliographystyle{splncs04}
\bibliography{egbib}

\appendix

\begin{algorithm}[t!]
		\caption{Goldilocks-curriculum Domain Randomization}
 		\label{GDR_algo}
 		\begin{algorithmic}[1]
		\REQUIRE $\Phi:$ simulation parameters space, $V:$ performance metric, $T:$ total timestep, $k:$ target performance score, $s:$ simulator,  $\theta^0:$ initial parameter of the model, $a:$ acquisition function in GP,  $n_{init}:$ the number of initial seeds in GP, $BO\_iter:$ the number of iterations in BO
		\ENSURE $\theta^T:$ the parameter of the model trained based on the curriculum.
		\FOR{$t \in \{1,2,\cdots, T\}$}
		\STATE $D \leftarrow \, \{ \}$
		\STATE \textit{//prepare $n_{init}$ initial seeds in BO}
		\FOR{$i \in \{1, 2, \cdots, n_{init}\}$}
		\STATE \textit{//sample simulation parameters}
		\STATE $\phi_{i} \sim \Phi$ 
		\STATE \textit{//generate synthetic data using the simulation parameters}
		\STATE $S_{\phi_{i}} \leftarrow s(\phi_{i})$ 
		\STATE \textit{//calculate the performance score and the distance from the target respectively}
		\STATE $V_{i} \leftarrow V(\theta^{t-1}, S_{\phi_{i}})$
		\STATE $d_{i} \leftarrow \, - | V_{i} - k|$
		\STATE $D \leftarrow \, D \bigcup \{ ( \phi_i, d_i) \}$
		
		\ENDFOR
		
		\STATE \textit{//run GP}
		\FOR{$i \in \{n_{init} + 1, \cdots, BO\_iter\}$}
		\STATE $\phi_i \leftarrow \argmax_{\phi \in \Phi} a(\phi, D)$
		
		\STATE $S_{\phi_{i}} \leftarrow s(\phi_{i})$ 
		\STATE $V_{i} \leftarrow V(\theta^{t-1}, S_{\phi_{i}})$
		\STATE $d_{i} \leftarrow \, - | V_{i} - k|$
		\STATE $D \leftarrow \, D \bigcup \{ ( \phi_i, d_i) \}$
		
		\ENDFOR
		
		\STATE \textit{//choose the optimal simulation parameters among the explored candidates}
		\STATE $j \leftarrow \argmax_{j \in \{1, 2,  \cdots, BO\_iter\}} d_j$
		\STATE $\phi^t \leftarrow \phi_j$
		
		\STATE \textit{//update the model parameter by training on the synthetic data}
		\STATE \textit{//One may include the previous data, equally}
		\STATE $\theta^t \leftarrow model\_update(\theta^{t-1}, S_{\phi^t}, S_{\phi^{1:t-1}})$
		
		\ENDFOR
\end{algorithmic}
\end{algorithm}

\begin{figure}[t!]
\centering
$
\begin{array}{ccc}
\includegraphics[width=7.5cm]{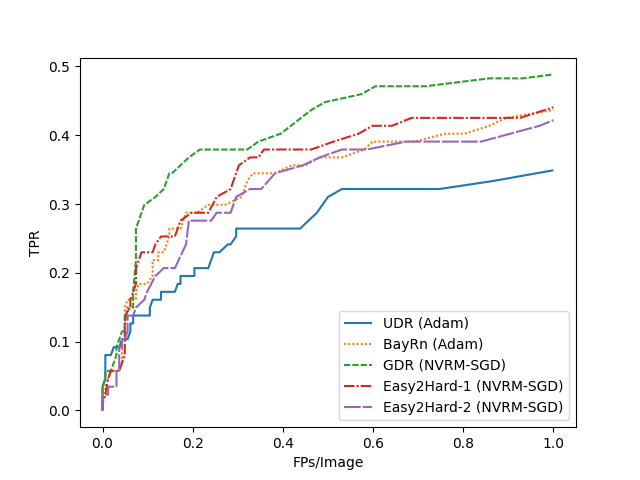}&
\includegraphics[width=7.5cm]{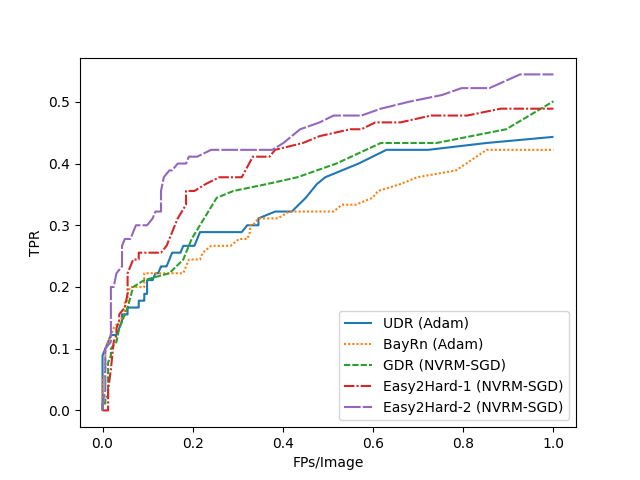}
\\
({\rm a})~{\rm Fold\,1} 
&
({\rm b})~{\rm Fold\,2} 
\\
\includegraphics[width=7.5cm]{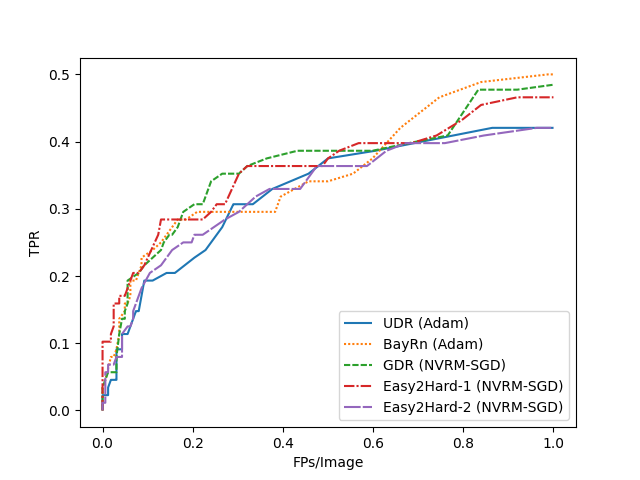}&
\includegraphics[width=7.5cm]{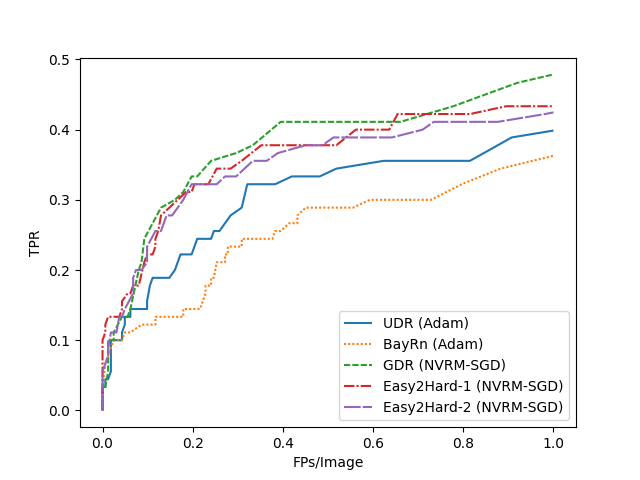}
\\
({\rm c})~{\rm Fold\,3} 
&
({\rm d})~{\rm Fold\,4} 
\\
\includegraphics[width=7.5cm]{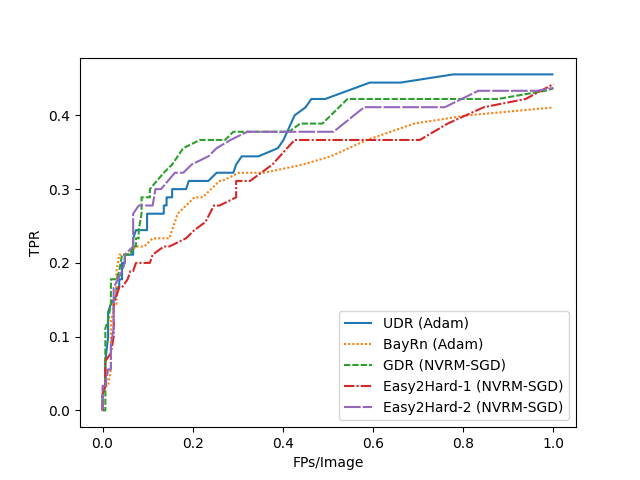}&
\\
({\rm e})~{\rm Fold\,5} 
&
\end{array}
$
\caption{FROC curves of different methods on $101$ pneumonia dataset.} 
\label{FROC}
\end{figure}

\section{Details of Simulator Construction}
\label{simulator_detail}

With a sophisticated configuration of a simulator, it is natural to expect that the machine learning model trained on synthetic data has the ability to detect lesions in medical images. However, it is obvious that what kind of simulator should be used must depend on what kind of cases we deal with. In this paper, we aim to detect small and difficult-to identify pneumonia lesions that might be overlooked by inexperienced radiologists. For those cases, we propose to exploit fractal Perlin noise to simulate pneumonia because the appearance of the noise might be able to capture the characteristics of pneumonia lesions well. Details of the simulator components are described in the following subsections. 

\subsection{Simulation of pneumonia lesions}
\label{smooth}

After generating a fractal Perlin noise image, we randomly transform its shape to create various shape patterns. The more detailed procedure is as follows. First, we create a circular mask whose pixel values inside the circle are normalized between 0 to 1. Next, we apply an affine transformation to randomly transform its shape. Then, we reshape a fractal Perlin noise image to the size of the deformed circle. Finally, we crop the reshaped noise image along the deformed shape of the circle. Through this process, we can provide a large variety of shape patterns of pseudo pneumonia lesions. 

There are several notable points in the above procedure. First, since the fractal Perlin noise image is reshaped to the size of the deformed mask, the size of the pseudo lesion is controlled by the radius $r$ of an initial circular mask. Thus the radius $r$ serves as another parameter for the simulation of pneumonia lesions. Second, we smooth the areas around the edge of an initial circle in order to generate more realistic pneumonia images when we insert the pseudo lesions. In other words, the pixel values near the boundary of the circle are adjusted to be smaller. Therefore, when cropping a fractal Perlin noise, by multiplying the pixel value at the deformed circle with the corresponding pixel value at the reshaped noise, the pixel values near the boundary of the cropped fractal Perlin noise tend to be smaller. We perform the smoothing by linearly decreasing the pixel values near the boundary. In more details, let $\alpha$ be the smoothing parameter, which takes a value between 0 and 1. Also, for each point in the initial circle, denote by $d$ its distance from the center of the circle. Then, if the distance $d$ is greater than $r (1-\alpha)$, the point is considered to be near the boundary of the circle, and thus the corresponding pixel value is decreased by multiplying $\frac{r-d}{r \cdot \alpha}$. This smoothing parameter $\alpha$ also serves as another parameter for the simulation of pneumonia lesions.

\subsection{Algorithm to determine the location of pneumonia lesion}

The generated pseudo pneumonia lesions are inserted into normal lung images to generate abnormal images, i.e., images with pneumonia lesions. In this process, in order to generate realistic abnormal images, we have to carefully determine where and how to insert the pneumonia lesions. 

With regard to determining where to superimpose the lesion, one solution would be the manual selection by experts \cite{DLA}. However, since we generate a large number of abnormal images through simulator, we have to automate the process of determining the insertion location. 

To begin with, it is natural and reasonable to think that the lesions should be inside the area of lungs. Therefore, automatically identifying the lung regions is beneficial to determine the lesion location, yet it is not obviously easy. Instead, we propose an alternative approach based on the observation that the pixel values at lung areas tend to be relatively small. In more detail, we apply the following algorithm until we find an appropriate location. First, once a normal lung image is given, a candidate location to insert a pneumonia lesion is randomly chosen from a specified range. Note that the background areas near the edges of the image are not appropriate as a insertion position; thus, we designate the location range in advance to avoid such a background area being chosen. Second, we calculate the average value of the pixels where the lesion will be inserted. Third, we judge whether the selected area is expected to be inside the lungs or not. Namely, if the average value is lower than a pre-defined threshold about whiteness, then use the area as a insertion location and terminates the algorithm. Otherwise, restart from the beginning; repeat this loop until a suitable location is found.

A practical challenge here is how to set the threshold value about whiteness. If the value is set high, a pseudo pneumonia lesion is more likely to be superimposed outside of lungs and an unnatural abnormal image will be generated. In order to avoid such a situation, the threshold value should be sufficiently small. However, when the value is set too small, an infinite loop might happen in some normal images. Considering the situation that some normal images look brighter and others look darker, and the overall pixel value differs from image to image, it is difficult and even unnecessary to fix the threshold value. For these reasons, we take a strategy of dynamically changing the threshold value if the loop repeats a certain number of times. More specifically, we set the initial value of the threshold about whiteness, and the value will be incremented if the iteration to find a insertion location repeats more than a maximum iteration times. Through this adaptive strategy, the termination of the algorithm is guaranteed. See the algorithm \ref{fig: algo} for details. 

In the algorithm \ref{fig: algo}, $M$ means the margins that should be ignored when a candidate of the insertion location is determined. In our experiment, for instance, the width and height of each normal X-ray image are respectively 1024 pixels, and we set $M$ to 240. Also, we fixed the maximum iteration times to 20, and we set the initial value of threshold about whiteness to 90 in our experiment.

\begin{algorithm}[t]
		\caption{Lesion Location Determination}
 		\label{fig: algo}
 		\begin{algorithmic}[1]
		\REQUIRE $I:$ a normal lung image, $L:$ a pseudo pneumonia lesion, $M:$ margin
		\ENSURE $x,y:$ the center point of the location to insert the lesion
		\STATE $width,\, height \leftarrow I.width, \, I.height$
		\STATE $lesion\_width, \,lesion\_height \leftarrow L.width, \, L.height$
		\STATE $iter\_cnt \leftarrow 0$
		\WHILE{True}
		\STATE $x \sim U(M, width-M)$ 	
		\STATE $y \sim U(M, height-M)$ 
		\STATE \textit{//calculate the average value of the pixels  where the lesion is inserted}
		\STATE $n \leftarrow 0, \, v \leftarrow 0$ 
		\FOR{$i \in \{1,2,\cdots, lesion\_width\}$}
		\FOR{$j \in \{1,2,\cdots, lesion\_height\}$}	
		\STATE \textit{//if the pixel is a part of the lesion}
		\IF{$L[i][j]>0$} 
		\STATE \textit{//add the corresponding pixel value of the normal lung image}
		\STATE $v \leftarrow v + I[x-lesion\_width/2+i][y-lesion\_height/2+j]$
		\STATE $n \leftarrow n+1$
		\ENDIF
		\ENDFOR
		\ENDFOR
        \STATE \textit{//judge whether the average pixel value is too white or not}
		\IF{$v/n \leq max\_acceptable\_whiteness$}
		\STATE {return $x,y$}
		\ELSE 
		\STATE $iter\_cnt \leftarrow iter\_cnt+1$
		\IF{$iter\_cnt > max\_iteration$}
		\STATE $max\_acceptable\_whiteness \leftarrow max\_acceptable\_whiteness + 1$
		\STATE $iter\_cnt \leftarrow 0$
		\ENDIF
		\ENDIF
		\ENDWHILE
		\end{algorithmic}
\end{algorithm}

\subsection{Insertion method}
\label{white}

After determining the location to insert a pneumonia lesion, we superimpose the lesion on the normal lung image. We have to carefully design how to insert the lesion in order to generate natural abnormal images. Here, we propose the insertion method using the Beer-Lambert law \cite{BeerLambert}, which is based on the principle of X-rays imaging. According to the Beer-Lambert law, the intensity of the X-ray radiation is attenuated by passing through a subject as follows:
\begin{align}
\label{formula: decrease}I = I_0 \, \exp \left(-\int \mu \, dl \right),    
\end{align}
where $I_0$ is an initial intensity of the X-ray radiation, $I$ 
is the intensity after the transmission through the subject, $\mu$ is an attenuation coefficient of the subject, and $dl$ represents the direction in which the X-ray radiation travels. The law indicates that, as the X-ray radiation passes through an object, it is attenuated and therefore the corresponding pixel values in the X-ray image become whiter. The Beer-Lambert law is useful for the simulation of X-ray imaging and used in previous studies \cite{BeerLambert,x-ray}. However, incorporating the law into our simulator is not straightforward, because we only have images of normal lungs and pneumonia lesions, and thus we have no information about attenuation coefficients and lengths of the objects in the images. Under these circumstances, we make the following two important assumptions to make use of the Beer-Lambert law in our simulator. 

First, we assume that there is a linear relationship between an intensity and a pixel value. This assumption seems to be reasonable to some extent, because an intensity and a pixel value are negatively correlated: the weaker the X-ray intensity is, the whiter the corresponding pixel and thus the larger its pixel value in the X-ray image will be. Also, when there is no object in the path of X-ray radiation, i.e., $I=I_0$, then the corresponding pixel value in the X-ray image becomes 0. On the other hand, when the radiation is completely blocked, i.e., $I=0$, then the corresponding pixel value will take the maximum value $v_{max}$, e.g., 255 under the 8-bit imaging representation. Based on these facts, we assume that the X-ray intensity and the corresponding pixel value of the image are linked by the linear function connecting the two endpoints: $(I_0, 0)$ and $(0, v_{max})$. By this assumption, we can freely transform the intensity of the X-ray radiation and the pixel value in the image.

Second, we assume that the images of pseudo pneumonia lesions are generated by X-ray imaging. By these assumptions, we can apply the Beer-Lambert law not only to normal lung images but also to pseudo pneumonia lesions images. 

Given these assumptions, each pixel value in the abnormal image, i.e., the image after inserting the lesion into the normal lung image, can be calculated in the following way. First, let $\mu_{in}$ be an attenuation coefficient, $v_{in}$ be a pixel value, and $I_{in}$ be the corresponding X-ray intensity of the normal image. Then, by the first assumption and the Beer-Lambert law, the following equation holds:
\begin{align*}
    \exp \left(-\int \mu_{in} \, dl \right) = \frac{I_{in}}{I_0} = \frac{v_{max} - v_{in}}{v_{max}}.  
\end{align*}
Here, when the maximum pixel value that the X-ray image can take is 255,
\begin{align}
\label{in}
    \exp \left(-\int \mu_{in} \, dl \right) = \frac{255 - v_{in}}{255}.
\end{align}
Next, let $\mu_{noise}$, $v_{noise}$, and $I_{noise}$ be respectively an attenuation coefficient, a pixel value, and the corresponding X-ray intensity of the pseudo pneumonia lesion. Then, by the assumptions and the Beer-Lambert law, 
\begin{align*}
    \exp \left(-\int \mu_{noise} \, dl \right) = \frac{I_{noise}}{I_0} = \frac{v_{max} - v_{noise}}{v_{max}}
\end{align*}
holds. Therefore, when the pseudo pneumonia lesion image is normalized between 0 and 1, i.e., $0 \leq v_{noise} \leq 1$ holds, then $v_{max}$ corresponds to 1 and thus 
\begin{align}
\label{noise}
    \exp \left(-\int \mu_{noise} \, dl \right) = 1 - v_{noise}.
\end{align}
Finally, by these equations \eqref{in} and \eqref{noise}, $I_{out}$, an intensity after inserting the lesion into the normal image can be expressed as follows:
\begin{align*}
    I_{out} &= I_0 \, \exp \left(-\int (\mu_{in} + \mu_{noise}) \, dl \right)\\
    &= I_0 \, \exp \left(-\int \mu_{in} \, dl \right) \cdot \exp \left(-\int \mu_{noise} \, dl \right) \\
    &= I_0 \cdot \frac{255 - v_{in}}{255} \cdot (1 - v_{noise}).
\end{align*}
Therefore, its corresponding pixel value $v_{out}$ can be calculated as
\begin{align}
    v_{out} = 255 \cdot \left(1-\frac{I_{out}}{I_0} \right) &= 255 \cdot \left(1- \frac{255 - v_{in}}{255} \cdot (1 - v_{noise})  \right)\\
\label{insertion}    &= v_{in}(1 - v_{noise}) + 255 \cdot v_{noise}.
\end{align}
From the equation \eqref{insertion}, each pixel value in the abnormal image is calculated by the corresponding pixel values in the normal image and in the pseudo pneumonia lesion. Moreover, considering that the lesion is normalized, i.e., $0 \leq v_{noise} \leq 1$, the equation \eqref{insertion} takes the form of interpolation, and the following requirements are naturally satisfied: (i) the pixel value after insertion never becomes darker. Especially when the pixel is within the insertion location, the corresponding pixel value becomes whiter than before insertion. (ii) the pixel value after insertion never becomes larger than the maximum pixel value, i.e., 255. (iii) the whiter the lesion is, the whiter the corresponding pixel values after insertion will be.

\begin{figure}[t]
 \begin{minipage}{0.32\hsize}
   \includegraphics[width=40mm]{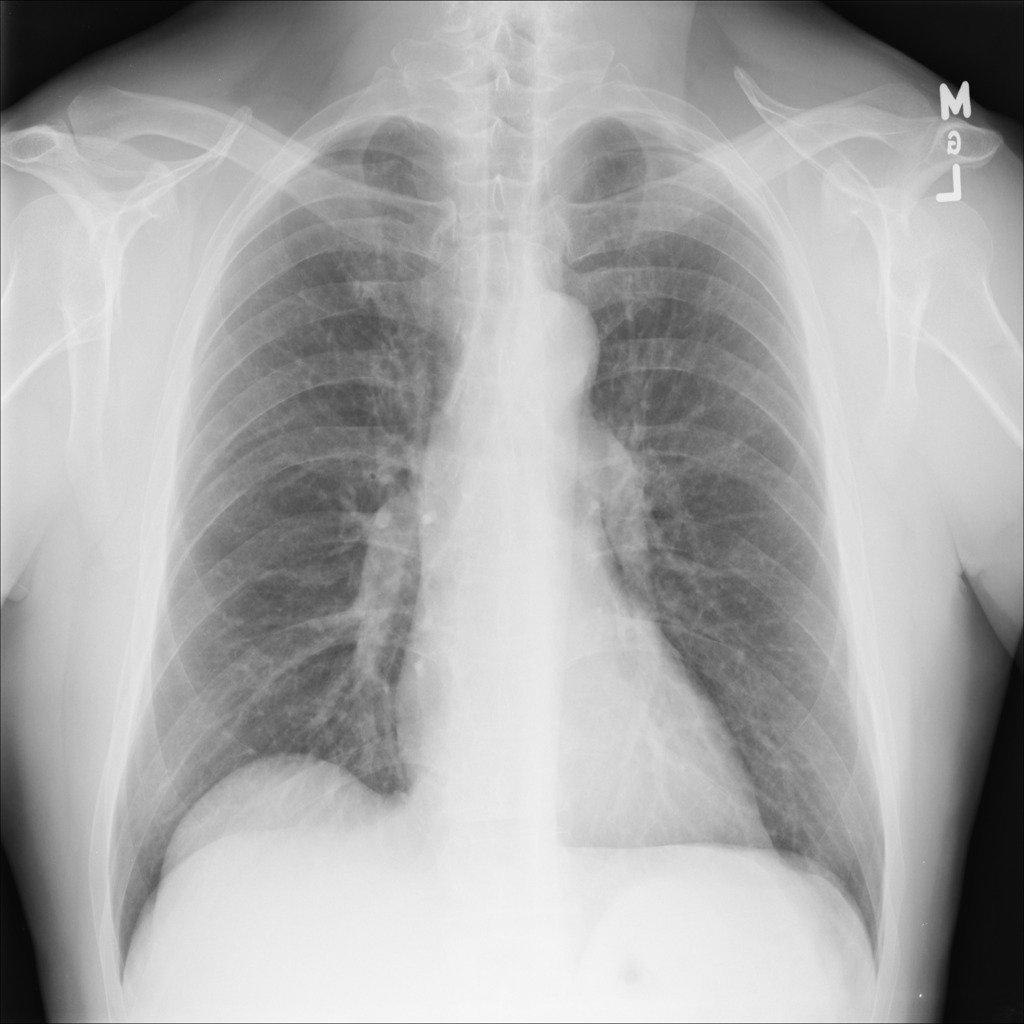}
 \end{minipage}
 \begin{minipage}{0.32\hsize}
  \includegraphics[width=40mm]{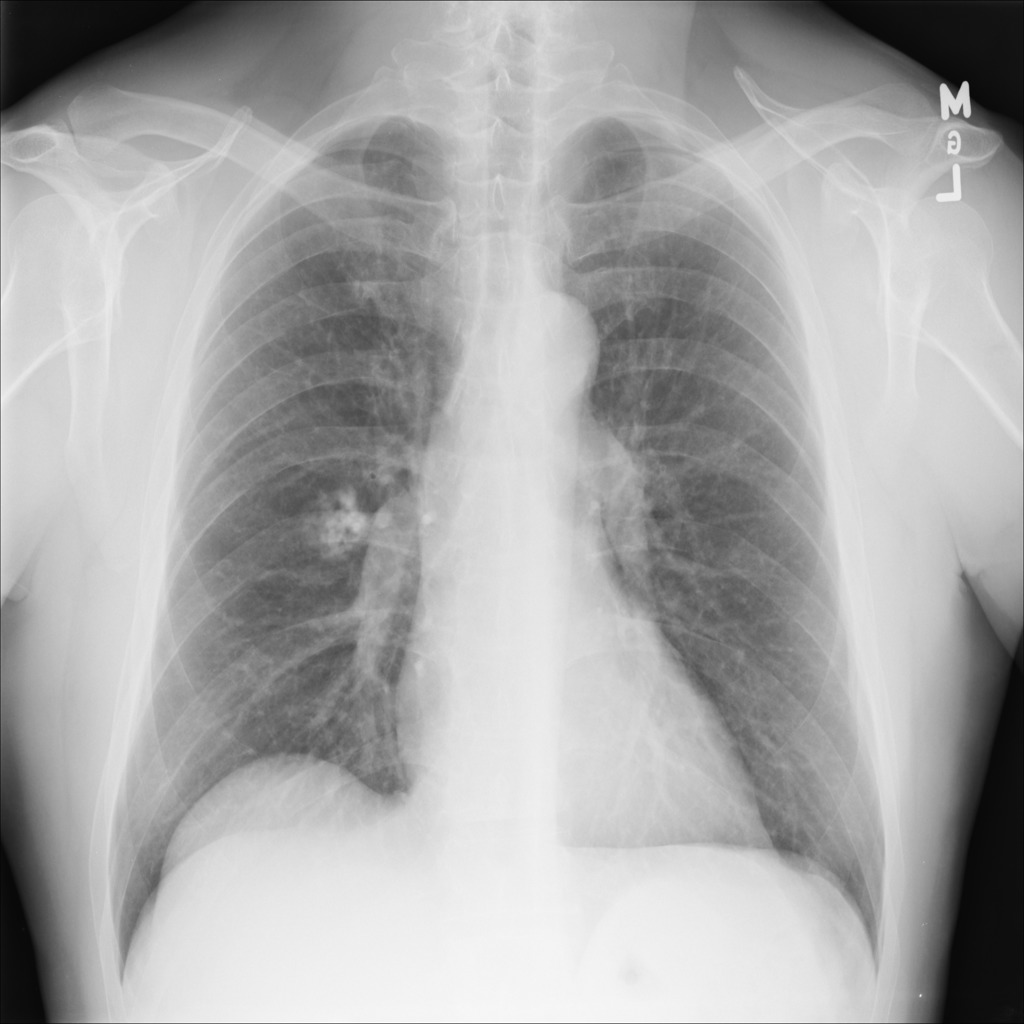}
 \end{minipage}
 \begin{minipage}{0.32\hsize}
  \includegraphics[width=40mm]{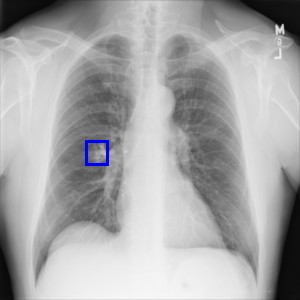}
 \end{minipage}
 \caption{Examples of abnormal images generated by the proposed simulator. The left is a normal lung image, and the middle is the abnormal image generated by the simulator. In the right, the location where the lesion has been inserted in the middle image is visualized by the blue bounding box.}
\label{figure3}
\end{figure}

Here, there are several points that should be noted. First, as already mentioned in the previous subsection, the areas around the border of the lesion have been smoothed by introducing the smoothing parameter $\alpha$. Second, it is considered necessary to adjust the scale of $v_{noise}$ appropriately in order to generate natural abnormal images. This is because, when the value $v_{noise}$ takes 1 in the original expression \eqref{insertion}, the corresponding pixel value $v_{out}$ gets 255; thus the pixel might be too white and the generated abnormal image will look unnatural. In order to prevent from generating such an unnatural abnormal image, we introduce the whiteness parameter $\beta$, which takes a value between 0 and 1. Then, each pixel value in the pseudo lesion is multiplied by $\beta$. Namely, after taking the whiteness parameter into account, each pixel value in the abnormal image can be calculated as follows: 
\begin{align}
    v_{out} = v_{in}(1 - \beta \cdot v_{noise}) + 255 \cdot \beta \cdot v_{noise}.
\end{align}
Examples of abnormal images generated by the proposed simulator are shown in the Figure \ref{figure3}.

\end{document}